\documentclass[a4paper,conference]{IEEEtran}

\usepackage{graphicx}
\usepackage{amsmath}
\usepackage{amssymb}

\usepackage[numbers,square,sort&compress]{natbib}

\usepackage{xspace}
\newcommand*{\eg}{{\it e.g.}\@\xspace}
\newcommand*{\ie}{{\it i.e.}\@\xspace}
\newcommand*{\etal}{{\it et~al.}\@\xspace}
\newcommand*{\cf}{{\it cf.}\@\xspace}

\usepackage{bm}
\newcommand{\mathbold}[1]{\bm{#1}}
\newcommand{\mbf}[1]{\mathbf{#1}}

\newcommand{\T}{\top}

\newcommand{\R}{\mathbb{R}}

\DeclareMathOperator{\diag}{diag}

\newcommand{\vomega}[0]{\mathbold{\omega}}

\renewcommand{\mid}[0]{\,|\,}

\newcommand{\vp}{\mbf{p}}
\newcommand{\vq}{\mbf{q}}
\newcommand{\vr}{\mbf{r}}

\newcommand{\vx}{\mbf{x}}
\newcommand{\vy}{\mbf{y}}
\newcommand{\vz}{\mbf{z}}

\newcommand{\MC}{\mbf{C}}
\newcommand{\MD}{\mbf{D}}

\newcommand{\MI}{\mbf{I}}

\newcommand{\MR}{\mbf{R}}

\newcommand{\MY}{\mbf{Y}}
\newcommand{\MZ}{\mbf{Z}}

\usepackage{subcaption}
\usepackage{multirow}

\usepackage{tikz,pgfplots}

\pgfplotsset{/pgf/number format/.cd, 1000 sep={}}

\pgfplotsset{every axis/.append style={
  grid style={line width=0.6pt,dotted,gray}}}

\pgfplotsset{every axis/.append style={
  legend style={inner xsep=1pt, inner ysep=0.5pt, nodes={inner sep=1pt, text depth=0.1em},draw=none,fill=none}
}}

\pgfplotsset{every axis/.append style={
  colorbar style={width=3mm,xshift=-2mm,major tick length=2pt}
}}

\definecolor{mycolor0}{rgb}{0.2667,0.4471,0.7098}
\definecolor{mycolor1}{rgb}{0.1647,0.6706,0.3804}
\definecolor{mycolor2}{rgb}{0.8275,0.2627,0.3059}
\definecolor{mycolor3}{rgb}{0.5216,0.4392,0.7176}
\definecolor{mycolor4}{rgb}{0.8118,0.7255,0.4118}
\definecolor{mycolor5}{rgb}{0.2745,0.7176,0.8157}

\definecolor{mylcolor0}{rgb}{0.6902,0.7686,0.8863}
\definecolor{mylcolor1}{rgb}{0.5451,0.8902,0.6941}
\definecolor{mylcolor2}{rgb}{0.9412,0.7490,0.7647}
\definecolor{mylcolor3}{rgb}{0.8627,0.8392,0.9176}
\definecolor{mylcolor4}{rgb}{0.9569,0.9373,0.8667}
\definecolor{mylcolor5}{rgb}{0.7529,0.9020,0.9373}
\definecolor{mylcolor6}{rgb}{0.8750,0.8750,0.8750}

\usetikzlibrary{plotmarks,shapes,arrows,positioning,3d}
\usetikzlibrary{arrows.meta}
\usetikzlibrary{decorations.text}
\pgfplotsset{compat=newest}

\newlength{\figurewidth}
\newlength{\figureheight}

\usepackage[pagebackref=true,breaklinks=true,letterpaper=true,colorlinks,bookmarks=false]{hyperref}
\usepackage[capitalize, nameinlink]{cleveref}
\crefname{section}{Sec.}{Secs.}

\usepackage{booktabs}
\usepackage{multirow}
\usepackage{tabularx}
\usepackage{longtable}

\author{\IEEEauthorblockN{Yuxin Hou}
\IEEEauthorblockA{Aalto University\\
Finland\\
yuxin.hou@aalto.fi}
\and
\IEEEauthorblockN{Muhammad Kamran Janjua}
\IEEEauthorblockA{National University of \\ Sciences and Technology,
Pakistan \\
mjanjua.bscs16seecs@seecs.edu.pk}
\and
\IEEEauthorblockN{Juho Kannala}
\IEEEauthorblockA{Aalto University\\
Finland\\
juho.kannala@aalto.fi}
\and
\IEEEauthorblockN{Arno Solin}
\IEEEauthorblockA{Aalto University\\
Finland\\
arno.solin@aalto.fi}}

\begin{document}

\title{Movement-induced Priors for Deep Stereo}

\maketitle

\IEEEpeerreviewmaketitle

\begin{abstract}
We propose a method for fusing stereo disparity estimation with movement-induced prior information. Instead of independent inference frame-by-frame, we formulate the problem as a non-parametric learning task in terms of a temporal Gaussian process prior with a movement-driven kernel for inter-frame reasoning. We present a hierarchy of three Gaussian process kernels depending on the availability of motion information, where our main focus is on a new gyroscope-driven kernel for handheld devices with low-quality MEMS sensors, thus also relaxing the requirement of having full 6D camera poses available. We show how our method can be combined with two state-of-the-art deep stereo methods. The method either work in a plug-and-play fashion with pre-trained deep stereo networks, or further improved by jointly training the kernels together with encoder--decoder architectures, leading to consistent improvement.

\end{abstract}

\section{Introduction}
\label{sec:intro}
\begin{figure}[t!]
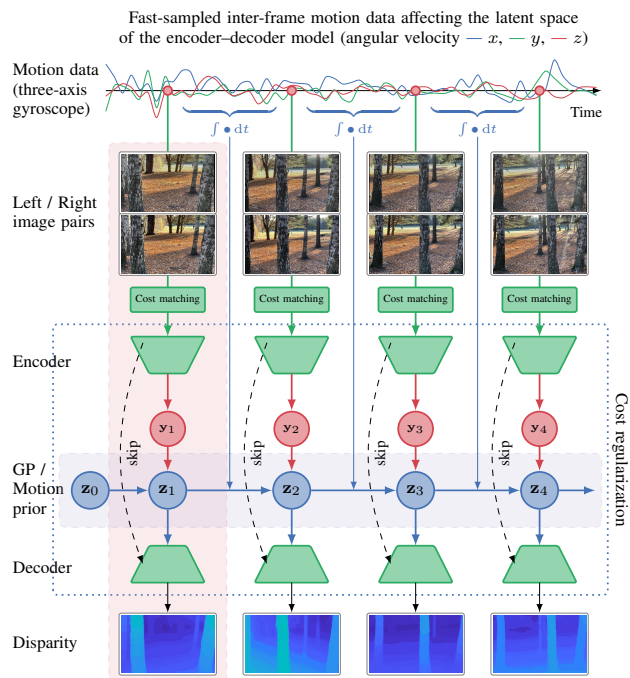

  \vspace*{-1em}
  \centering
  \resizebox{1.03\columnwidth}{!}{%
  \footnotesize 
  \begin{tikzpicture}

    \newcommand{\drawnet}[3]{%
      \tikzstyle{block} = [rounded corners=0.5pt,minimum width=1mm,minimum height=1mm,inner sep=0,draw=#3!50!black,fill=#3]
      \foreach \j in {0,...,4} 
        \node[block,minimum height={0.25cm+0.2cm*\j}] at (#1-.1*\j,#2) {};  
      \foreach \j in {0,...,4} 
        \node[block,minimum height={0.25cm+0.2cm*\j}] at (#1+.1*\j,#2) {};}

    \newcommand{\encoder}[3]{%
      \tikzstyle{block} = [rounded corners=0.5pt,minimum width=1mm,minimum height=1mm,inner sep=0,draw=#3!50!black,fill=#3]
      \foreach \j in {0,...,4} 
        \node[block,minimum width={0.25cm+0.2cm*\j}] at (#1,#2+.1*\j) {};} 

    \newcommand{\decoder}[3]{%
      \tikzstyle{block} = [rounded corners=0.5pt,minimum width=1mm,minimum height=1mm,inner sep=0,draw=#3!50!black,fill=#3]
      \foreach \j in {0,...,4} 
        \node[block,minimum width={0.25cm+0.2cm*\j}] at (#1,#2-.1*\j) {};} 

    \newcommand{\costvol}[2]{%
      \foreach \y in {1,...,6} {
        \draw[draw=mycolor1,fill=mycolor1!10,rounded corners=1pt] ({-0.75+0.05*\y+2*#1}, {#2+.07*\y}) -- ++({1.5-.1*\y},0) -- ++({-.25+.03*\y},{0.5-0.05*\y}) -- ++({-1+0.05*\y},0) -- cycle;}}
        
    \renewcommand{\encoder}[3]{%
      \node at (#1,#2) {
      \tikz\draw[rounded corners=1pt,inner sep=0,draw=#3,fill=#3!50,thick] (-6mm,3mm) -- (6mm,3mm) -- (3mm,-3mm) -- (-3mm,-3mm) -- cycle;};
    }    
     
    \renewcommand{\decoder}[3]{%
      \node at (#1,#2) {
      \tikz\draw[rounded corners=1pt,inner sep=0,draw=#3,fill=#3!50,thick] (-3mm,3mm) -- (3mm,3mm) -- (6mm,-3mm) -- (-6mm,-3mm) -- cycle;};
    }

    \draw[mycolor0,dotted,rounded corners=2pt,thick] (1.5mm, -17mm) rectangle ++(89.5mm,44mm);  
    \node[rotate=270] at (9.3,0.5) {Cost regularization};

    \tikzstyle{arrow} = [draw=black!10, single arrow, minimum height=10mm, minimum width=3mm, single arrow head extend=1mm, fill=black!10, anchor=center, rotate=-90, inner sep=2pt]

    \node[text width=2cm, align=left] at (0.5,6.5) {Motion data \mbox{(three-axis} \mbox{gyroscope)}};    
    \node[text width=2cm, align=left] at (0.5,4.5) {Left / Right \mbox{image pairs}};
    \node[text width=2cm, align=left] at (0.5,2.1) {Encoder};
    \node[text width=2cm, align=left] at (0.5,0) {GP / \\ {Motion} \\ prior};
    \node[text width=2cm, align=left] at (0.5,-1.25) {Decoder};
    \node[text width=2cm, align=left] at (0.5,-2.5) {Disparity};
    \node[text width=8cm, align=center] at (5,7.5) {Fast-sampled inter-frame motion data affecting the latent space of the encoder--decoder model \mbox{(angular velocity 
    \tikz[baseline]\draw[draw=mycolor0,yshift=2pt] (0,0)--(.25,0); $x$,  
    \tikz[baseline]\draw[draw=mycolor1,yshift=2pt] (0,0)--(.25,0); $y$,  
    \tikz[baseline]\draw[draw=mycolor2,yshift=2pt] (0,0)--(.25,0); $z$)}};

    \node[text width=2cm, align=left] at (9.5,6.15) {\scriptsize Time};

    \draw[draw=mycolor2,dashed,fill=mycolor2!50,rounded corners=1mm, opacity=.2] (1.05,5.65) rectangle (2.95,-3.15);
    \draw[draw=mycolor3,dashed,fill=mycolor3!50,rounded corners=1mm, opacity=.2] (.25,.6) rectangle (9,-.6);

    \draw[black,-latex] (1.0,6.5) -- (9,6.5);     
    \draw[mycolor0] plot [smooth] coordinates{ (1.000,6.931) (1.137,6.727) (1.273,6.462) (1.410,6.750) (1.546,6.518) (1.683,6.608) (1.819,6.329) (1.956,6.639) (2.092,6.649) (2.229,6.824) (2.365,6.806) (2.502,6.502) (2.638,6.657) (2.775,6.725) (2.911,6.585) (3.048,6.464) (3.184,6.620) (3.321,6.641) (3.457,6.455) (3.594,6.694) (3.730,6.507) (3.867,6.660) (4.003,6.629) (4.140,6.617) (4.276,6.595) (4.413,6.462) (4.549,6.718) (4.686,6.700) (4.823,6.607) (4.959,6.631) (5.096,6.670) (5.232,6.568) (5.369,6.565) (5.505,6.529) (5.642,6.431) (5.778,6.581) (5.915,6.427) (6.051,6.436) (6.188,6.450) (6.324,6.501) (6.461,6.536) (6.597,6.494) (6.734,6.686) (6.870,6.555) (7.007,6.755) (7.143,6.476) (7.280,6.432) (7.416,6.356) (7.553,6.325) (7.689,6.656) (7.826,6.394) (7.962,6.476) (8.099,6.529) (8.235,6.997) (8.372,6.744) (8.509,6.598) (8.645,6.543) (8.782,6.487) (8.918,6.420) };
    \draw[mycolor1] plot [smooth] coordinates{ ((1.000,6.162) (1.137,6.577) (1.273,6.571) (1.410,6.495) (1.546,6.363) (1.683,6.826) (1.819,6.222) (1.956,6.729) (2.092,6.428) (2.229,6.477) (2.365,6.428) (2.502,6.458) (2.638,6.424) (2.775,6.508) (2.911,6.600) (3.048,6.414) (3.184,6.446) (3.321,6.528) (3.457,6.477) (3.594,6.285) (3.730,6.464) (3.867,6.356) (4.003,6.524) (4.140,6.535) (4.276,6.502) (4.413,6.534) (4.549,6.340) (4.686,6.514) (4.823,6.363) (4.959,6.324) (5.096,6.578) (5.232,6.604) (5.369,6.486) (5.505,6.624) (5.642,6.562) (5.778,6.365) (5.915,6.499) (6.051,6.488) (6.188,6.459) (6.324,6.460) (6.461,6.453) (6.597,6.494) (6.734,6.499) (6.870,6.392) (7.007,6.554) (7.143,6.644) (7.280,6.479) (7.416,6.430) (7.553,6.379) (7.689,6.335) (7.826,6.447) (7.962,6.648) (8.099,6.818) (8.235,6.704) (8.372,6.578) (8.509,6.487) (8.645,6.500) (8.782,6.490) (8.918,6.428) };
    \draw[mycolor2] plot [smooth] coordinates{ (1.000,6.409) (1.137,6.519) (1.273,6.613) (1.410,6.554) (1.546,6.399) (1.683,6.468) (1.819,6.062) (1.956,6.477) (2.092,6.438) (2.229,6.502) (2.365,6.396) (2.502,6.610) (2.638,6.730) (2.775,6.679) (2.911,6.519) (3.048,6.420) (3.184,6.450) (3.321,6.484) (3.457,6.285) (3.594,6.395) (3.730,6.591) (3.867,6.539) (4.003,6.585) (4.140,6.563) (4.276,6.463) (4.413,6.428) (4.549,6.371) (4.686,6.500) (4.823,6.441) (4.959,6.482) (5.096,6.567) (5.232,6.526) (5.369,6.479) (5.505,6.576) (5.642,6.546) (5.778,6.423) (5.915,6.475) (6.051,6.556) (6.188,6.556) (6.324,6.484) (6.461,6.520) (6.597,6.556) (6.734,6.446) (6.870,6.431) (7.007,6.531) (7.143,6.630) (7.280,6.587) (7.416,6.568) (7.553,6.456) (7.689,6.375) (7.826,6.362) (7.962,6.391) (8.099,6.509) (8.235,6.405) (8.372,6.382) (8.509,6.452) (8.645,6.551) (8.782,6.562) (8.918,6.551) };

    \node[shape=circle,draw=mycolor0,fill=mycolor0!50,thick,minimum width=2em] (z0) at (0.75,0) {$\vz_0$};

    \foreach \x in {1,...,4} {

      \draw[thick, draw=mycolor1,-latex] ({2*\x},{6.5}) -- ({2*\x},{2.5});

      \node[shape=circle,draw=mycolor2,fill=mycolor2!50,thick,minimum width=4pt,inner sep=0] at ({2*\x},6.5) {};

      \node[shape=rectangle,draw=black!50,minimum width=1.5cm,minimum height=1cm, rounded corners=1pt, inner sep=1pt,fill=white] (I\x) at (2*\x,4) {\includegraphics[width=1.5cm]{fig/sketch/left-\x}};
      \node[shape=rectangle,draw=black!50,minimum width=1.5cm,minimum height=1cm, rounded corners=1pt, inner sep=1pt,fill=white] (I\x) at (2*\x,5) {\includegraphics[width=1.5cm]{fig/sketch/right-\x}};

      \encoder{{2*\x}}{2.2}{mycolor1}
      \node (e\x) at ({2*\x},2) {};

      \node[shape=circle,draw=mycolor2,fill=mycolor2!50,thick,minimum width=1.5em] (y\x) at ({2*\x},1) {\tiny $\vy_\x$};

      \draw[-latex,thick,mycolor2] (e\x)->(y\x);

      \node[shape=circle,draw=mycolor0,fill=mycolor0!50,thick,minimum width=2em] (z\x) at ({2*\x},0) {$\vz_\x$};

      \draw[-latex,thick,mycolor2] (y\x)->(z\x);

      \node[rounded corners=1pt,inner sep=0,draw=mycolor1,fill=mycolor1!50, minimum width=12mm, minimum height=4mm, thick] at ({2*\x},3.1) {\tiny Cost matching};

      \decoder{{2*\x}}{-1.2}{mycolor1}
      \node (d\x) at ({2*\x},-1) {};

      \draw[-latex,dashed] ({2*\x-0.4},2.4) to [bend right=20] node[below,rotate=90] {\scriptsize skip} ++(0,-3.6);

      \draw[-latex,thick,mycolor0] (z\x)->(d\x);

      \node[shape=rectangle,draw=black!50,minimum width=1.5cm,minimum height=1cm, rounded corners=1pt,inner sep=1pt,fill=white] (disp\x) at (2*\x,-2.5) {\includegraphics[width=1.5cm]{fig/sketch/disp-\x}};

      \draw[-latex] (d\x)++(0,-.5)->(disp\x);

    }

    \foreach \x in {1,...,3} {

      \node[] at ({2*\x+1},{6.0}) {\textcolor{mycolor0}{$\underbrace{\hspace{1.5cm}}_{\int \bullet\,\mathrm{d}t}$}};

      \draw[mycolor0,-latex] ({2*\x+1},{5.75}) -- ({2*\x+1},0);
      
    }

    \node (z5) at (9,0) {};

    \draw[-latex,draw=mycolor0,thick] (z0)->(z1);
    \draw[-latex,draw=mycolor0,thick] (z1)->(z2);
    \draw[-latex,draw=mycolor0,thick] (z2)->(z3);
    \draw[-latex,draw=mycolor0,thick] (z3)->(z4);
    \draw[-latex,draw=mycolor0,thick] (z4)->(z5);
  
  \end{tikzpicture}}

  \caption{Sketch of the logic of our approach: The inputs are a sequence of stereo image pairs and corresponding motion information (\eg, a stream of three-axis gyroscope measurements). We present a framework for fusing deep stereo inferences (denoted by vertical red block) with movement-induced coupling (see horizontal blue block). The GP inference (blue block) couples the latent-space encodings of the methods based on the movement-induced prior that carries over information between stereo pairs.}

  \label{fig:architecture}
  \vspace*{-1em}
\end{figure}

Stereo matching refers to the problem of estimating a dense disparity map between a rectified pair of images. Given the internal and external camera parameters for the two images (see \cref{fig:architecture}), a dense depth map can be obtained from the estimated disparities via triangulation. Depth estimation is essential in many computer vision applications like autonomous driving and 3D model reconstruction. 
However, searching the corresponding points in a stereo pair is an ill-posed problem in the general case, since a scene may contain multiple disjoint surfaces at different depths, while each camera senses only a single optical surface. There can be equivalent surfaces that produce the same images. Also, stereo matching can be particularly challenging with  texureless regions, repetitive patterns, and signal noise, which urges the need for good priors to regularize the solution.
During the recent years, deep learning has been shown to be effective in learning suitable priors and convolutional neural networks (CNNs) are currently the main paradigm for disparity estimation \cite{mayer2016large, zbontar2016stereo, kendall, chang2018pyramid, nie2019multi, zhang2019ga, guo2019group}.

So far, most binocular stereo matching methods have only considered single image pairs. Also, current popular benchmarks like KITTI-2012~\cite{geiger2012we}, KITTI-2015~\cite{menze2015object}, and Middlebury~v3 \cite{scharstein2014high} only focus on single-pair inference. However, in many real-world applications, the stereo cameras actually keep collecting sequences of image pairs, and the additional information from multiple pairs could potentially be useful for more stable and robust disparity estimation. Introducing information sharing between frames helps in obtaining temporal consistency. Also, estimation in occluded regions could benefit from using consecutive pairs. In addition, on small devices a fixed stereo camera is constrained to have a small baseline, but utilizing a sequence of image pairs could effectively allow to extend the baseline beyond the physical size of the device.

In this paper, we focus on solving disparity estimation for image pair sequences. We propose an intuitive scheme to introduce movement-induced prior information by framing the problem as a non-parametric Gaussian process (GP, see, \eg, \cite{Rasmussen+Williams:2006}) inference task. We fuse information between the latent feature representations produced for the stereo pairs by deep networks that have encoder--decoder structures. We present three different GP kernels depending on the availability of motion information: \emph{(a)}~a full pose kernel is used when full (6-dof) relative motion with rotations and translations is known, \emph{(b)}~a gyroscope kernel is used if only angular rates of the relative orientation changes are known (in a sequence temporally in-between the camera frames), and \emph{(c)}~a time-decay kernel is used in cases where motion is unknown (only relative time difference between frames used as input). In many cases, full relative motion could be available through the use of (visual-inertial) odometry technology, which may be available in moving vehicles for navigation purposes, and currently even on smartphones (\eg, ARKit by Apple and ARCore by Google). On low-end smartphones and wearable devices, typically the 6-dof pose estimation is infeasible, but 3-axis gyroscope readings are available. The time-decay kernel can be used when no exogenous information is available.

It should be noted that we do not address dynamic objects via optical flow \cite{dosovitskiy2015flownet} or scene flow \cite{sceneflow}. Yet, when we evaluate our approach using the KITTI data set, which contains moving objects, the results show that our approach behaves robustly for dynamic scenes. In addition, our method is complementary to post-processing methods, such as depth map fusion \cite{merrell2007real}.

In summary, the contributions of this paper are as follows.
{\em (i)}~We present a hierarchy of three movement-induced Gaussian process (GP) kernels for improving disparity estimation from binocular image sequences through probabilistic information fusion---with focus on a novel lightweight gyroscope-kernel. {\em (ii)}~We present an approach for jointly training the GP hyperparameters with the deep disparity estimation networks. {\em (iii)}~We show that the proposed priors give consistent accuracy improvement, when combined with both DispNetC and PSMNet, which are recent architectures from two classes of CNN based stereo methods.

\section{Related work}
\label{sec:related}
The pipeline of conventional stereo matching methods generally consists of matching cost computation, cost aggregation, disparity optimization, and post-processing refinement \cite{scharstein2002taxonomy}.
Common metrics for computing matching cost are squared difference (SSD), sum of absolute differences (SAD), or normalized cross-correlation (NCC). To aggregate costs, there are local methods and global methods. Many local methods attempt to aggregate costs of neighboring pixels by defining local support windows \cite{bobick1999large, veksler2001stereo},
while global methods aim to minimize a global energy function \cite{boykov2001fast, kolmogorov2001computing, sun2003stereo, klaus2006segment}.
Semi-global matching (SGM, \cite{sgm}) approximate the global optimization by computing 1D constraints in many directions.
Though hand-crafted features and cost metrics are widely used, during recent years, learnt features and cost functions have been shown to be promising in stereo matching problems, where traditional methods face problems. Zbontar and LeCun~\cite{zbontar2015computing} showed that deep convolutional neural networks can be trained for matching image patches and works well with SGM. 

Inspired by the pipelines of traditional methods, most deep stereo methods also compute a cost volume by correlating features among the horizontal lines. According to the way of forming cost volume, most methods can be categorized into two classes. One class computes correlation scores between features, like DispNetC~\cite{mayer2016large} and its extensions \cite{pang2017cascade, liang2018learning}. Mayer \etal~\cite{mayer2016large} proposed DispNet, which predicts disparity maps directly via a simple encoder--decoder architecture with skip connections between contracting and expanding network parts. DispNetC is a variant of DispNet, which utilizes an explicit correlation layer \cite{dosovitskiy2015flownet} to obtain a 1D correlation map for each disparity level via inner product of feature vectors.
Following the idea of DispNetC, there are many extensions to it, like CLR \cite{pang2017cascade} and iResNet \cite{liang2018learning}. Some methods also integrate other tasks like semantic segmentation \cite{yang2018segstereo} to further improve the performance.

Another class of deep stereo methods concatenates features to build 4D cost volumes to incorporate context and geometry, as the single-channel correlation map of DispNetC decimates the feature dimension causing loss of information. Kendall \etal~\cite{kendall} proposed GC-Net, which concatenates unary features across each disparity level to form the cost volume and then uses 3D convolutions to learn to disparity prediction.
PSMNet~\cite{chang2018pyramid} utilizes a Spatial Pyramid Pooling (SPP) module before building cost volume to incorporate hierarchical representations and enlarge the receptive fields. To regularize the cost volume, PSMNet uses stacked hourglass architecture to learn more context information. Based on PSMNet, several recent works modified the scheme of cost matching computation to merge features \cite{guo2019group, nie2019multi}. GA-Net~\cite{zhang2019ga} proposed guided aggregation layers to regularize matching costs. Apart from improving the architecture of networks, \cite{poggi2019} introduces a new method to exploit additional sparse cues provided by LiDARs.

Most of the aforementioned works and also other recent deep stereo methods focus on new network architectures, either obtained through manual or automatic search \cite{autodispnet}. Another area of previous improvements has been self-supervised learning of deep stereo networks \cite{Zhong2017,Zhou2017}, and some of these methods have utilized binocular sequences for self-adapting the network weights after pre-training \cite{tonioni2019real}. Though they utilize stereo sequences during training and adaptation, the inference is still only based on information in single input image pairs.

In contrast, in this work, we focus on inter-frame inference, which aims to exploit the information between frames of a binocular stereo sequence in a computationally efficient manner by introducing a movement-induced prior. The closest related works are from monocular depth estimation, where temporal information fusion has been used before \cite{hou2019multi,Tananaev2018}, either assuming full knowledge of camera motion \cite{hou2019multi} or without motion information but with using recurrent networks to exploit temporal information explicitly \cite{Tananaev2018, zhong2018open}. Since using recurrent networks can introduce extra parameters and make training more difficult, \cite{hou2019multi} introduce a Gaussian process prior into depth prediction for monocular image sequences to make robust estimation.  The GP prior enables the model to fuse information from previous frames in the latent space efficiently and the features of the GP prior help avoid overfitting (see discussion in \cite{Rasmussen+Williams:2006}).

Probabilistic priors have been considered before with the help of Markov Random Field (MRF) and Conditional Random Fields (CRF) have long been used by traditional methods to impose global priors to optimize costs, which related to Gaussian process priors. \cite{zhang2007estimating} refomulated stereo matching as a MAP estimation problem for both the disparity map and the MRF parameters. The CRF approach was first proposed for building probabilistic models to segment and label sequence data \cite{lafferty2001conditional}. The work of \cite{scharstein2007learning} developed a novel CRF model for stereo estimation. \cite{knobelreiter2017end} proposed an end-to-end architecture that combines CNN and CRF for stereo estimation. However, both these probabilistic models with MRFs or CRFs only consider the sequences of pixels within single inputs, which is different from our prior that can introduce motion information for inter-frame fusion.

In contrast to these previous papers, we show that temporal fusion is useful also for learning-based stereo matching. Moreover, we present several movement-induced priors, including a new gyroscope-driven Gaussian process kernel that only requires rotational rate observations, which could be directly obtained from a gyroscope attached to the camera. That is, full odometry is not necessarily needed and this improves the applicability of the method to different use cases, where robust 6-dof camera tracking is either not available or infeasible.

\section{Methods}
\label{sec:methods}
Our approach is illustrated in \cref{fig:architecture}. The vertical information flow can be replaced by existing deep stereo matching networks, which predict a disparity map given rectified stereo pairs. The horizontal information flow is inferred with the help of a probabilistic prior. We explain the details of how to incorporate the two flows in \cref{sec:gp}. The movement-inducing pose and gyroscope kernels are then introduced in \cref{sec:movement-induced}.

\begin{figure}[t]
  \captionsetup{justification=centering}
  \begin{subfigure}{\columnwidth}
    \centering
    \input{./fig/path.tex}
    \caption{KITTI path with every 5\textsuperscript{th} pose drawn }
    \label{fig:path}  
  \end{subfigure}\\[3pt]
  \begin{subfigure}{.48\columnwidth}

    \setlength{\figurewidth}{.8\textwidth}
    \centering\scriptsize
    \begin{tikzpicture}
    \begin{axis}[%
      axis on top,xmin=1,xmax=144,ymin=1,ymax=144,
      xlabel={Frame \#},ylabel={Frame \#}, y dir=reverse,
      width=\figurewidth, height=\figurewidth,
      scale only axis,yticklabel style={rotate=90}
      ]
      \addplot [forget plot] graphics [xmin=1,xmax=144,ymin=1,ymax=144] {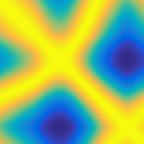};
    \end{axis}   
    \end{tikzpicture}     
    \caption{Covariance based on full gyro cross-distance (\cref{eq:fullgyrodist})}

    \label{fig:full-dist}  
  \end{subfigure}
  \hfill
  \begin{subfigure}{.48\columnwidth}

    \setlength{\figurewidth}{.8\textwidth}
    \centering\scriptsize
    \begin{tikzpicture}
    \begin{axis}[%
      axis on top,xmin=1,xmax=144,ymin=1,ymax=144,
      xlabel={Frame \#},ylabel={Frame \#}, y dir=reverse,
      width=\figurewidth, height=\figurewidth,
      scale only axis,yticklabel style={rotate=90}
      ]
      \addplot [forget plot] graphics [xmin=1,xmax=144,ymin=1,ymax=144] {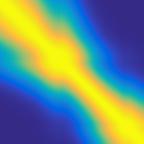};
    \end{axis}
    \end{tikzpicture} 
     \caption{Our Markovian gyro covariance based on kernel \cref{eq:gyrokernel}}
    \label{fig:markovian}      
  \end{subfigure}\\[3pt]
  \begin{subfigure}{.48\columnwidth}

    \setlength{\figurewidth}{.8\textwidth}
    \centering\scriptsize
    \begin{tikzpicture}
    \begin{axis}[%
      axis on top,xmin=1,xmax=144,ymin=1,ymax=144,
      xlabel={Frame \#},ylabel={Frame \#}, y dir=reverse,
      width=\figurewidth, height=\figurewidth,
      scale only axis,yticklabel style={rotate=90}
      ]
      \addplot [forget plot] graphics [xmin=1,xmax=144,ymin=1,ymax=144] {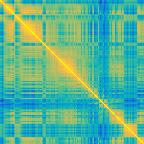};
    \end{axis}   
    \end{tikzpicture}     
    \caption{Latent $\vz$ correlation\\ (without prior)}
    \label{fig:without}  
  \end{subfigure}
  \hfill
  \begin{subfigure}{.48\columnwidth}

    \setlength{\figurewidth}{.8\textwidth}
    \centering\scriptsize
    \begin{tikzpicture}
    \begin{axis}[%
      axis on top,xmin=1,xmax=144,ymin=1,ymax=144,
      xlabel={Frame \#},ylabel={Frame \#}, y dir=reverse,
      width=\figurewidth, height=\figurewidth,
      scale only axis,yticklabel style={rotate=90}
      ]
      \addplot [forget plot] graphics [xmin=1,xmax=144,ymin=1,ymax=144] {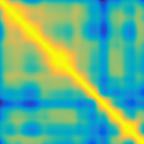};
    \end{axis}   
    \end{tikzpicture}     
    \caption{Latent $\vz$ correlation\\ (with prior)}
    \label{fig:with}      
  \end{subfigure}  
\caption{(a--c)~A KITTI path and the corresponding rotational gyro covariance and our Markovian pose-to-pose version of it which kills off spurious long-term correlation (same directions have short distance even if large translation). (d)~Sample correlation of latent space of pre-trained DispNetC between frames. The matrix is computed by flattening the latent representations of dimension $1024{\times}6{\times}20$. (e)~With the gyroscope-driven GP applied the sample covariance is less noisy and more structured.}
  \label{fig:comparison}
  \vspace*{-1em}
\end{figure}

\subsection{Probabilistic Gaussian process inference}
\label{sec:gp}
Most existing deep stereo methods can be categorized into two classes according to the way of forming cost volume. One class computes correlation scores between features, like DispNetC~\cite{mayer2016large} and its extensions \cite{pang2017cascade, liang2018learning}. Another class concatenates features to build 4D cost volumes, like GC-Net~\cite{kendall}, PSMNet~\cite{chang2018pyramid}, and their extensions \cite{guo2019group, nie2019multi, zhang2019ga}. Though the way of aggregation features can vary a lot from method to method, most of them share a common property: a fully-convolutional encoder--decoder architecture is widely used to regularize cost volumes---a property we tap into. As our proposed prior will not affect computation of cost volume and only incorporate with latent codes of the encoder--decoder part, it should be applicable to almost all existing learning-based methods. In this paper, we select the most representative two models of the two classes, DispNetC~\cite{mayer2016large} and PSMNet~\cite{chang2018pyramid}, to validate the effectiveness of our method. DispNetC uses a simple encoder--decoder architecture, while PSMNet uses a stacked hourglass module which consists of three coupled encoder--decoders for cost regularization, and we will apply the proposed prior on the first encoder--decoder only. 

Instead of dealing with each left--right pair independently, we introduce a probabilistic prior to the latent space of encoder--decoders, which will be spanned by a covariance function (kernel). To share the temporal information within the sequence, we modify the outputs $\vy_i$ from the encoder (see horizontal blue block in in \cref{fig:architecture}) by a Gaussian process (GP) regression model \cite{Rasmussen+Williams:2006}:
\begin{equation}
\begin{split}\label{eq:gp}
  z_j(t) &\sim \mathrm{GP}(0, \kappa(t,t')), \\
  y_{j,i} &= z_j(t_i) + \varepsilon_{j,i}, \quad  \varepsilon_{j,i} \sim \mathrm{N}(0, \sigma^2),
\end{split}
\end{equation}
where the top part denotes the GP prior over the latent encodings with covariance function $\kappa(\cdot,\cdot)$, and the bottom a Gaussian likelihood function. The $z_j(t_i)$ are the values of the latent function $\vz_i$ at time $t_i$, which will be fed into the decoder. The encoder output $\vy_i$ can now be seen as a `corrupted' version of the true (unknown) latent encodings $\vz_i$. The range of $j$ depends on the dimension of the output of the encoder. Specifically, for DispNetC the size will be $1024\times\frac{1}{64}H\times\frac{1}{64}W$, and for PSMNet the size will be $\frac{1}{16}D\times\frac{1}{16}H\times\frac{1}{16}W\times64$, where the maximum disparity $D$ is set to 192. We flatten the latent representations before the GP regression.

To obtain the $z_j(t)$ in \cref{eq:gp} (to feed the decoder), we solve independent GP regression problems for each dimension of the latent code with a single matrix inversion, as the posterior covariance is not dependent on any learnt representations of input images (\ie, $\vy$ does not appear in the posterior variance terms in \cref{eq:gp-regression}). The posterior mean and covariance will be given by \cite{Rasmussen+Williams:2006}:
\begin{equation}
\begin{split}\label{eq:gp-regression}
  \!\!\!\!\mathbb{E}[\MZ \mid \mathcal{D}] &{=} \MC \, (\MC+\sigma^2\,\MI)^{-1}\,\MY, \\
  \!\!\!\!\mathbb{V}[\MZ \mid \mathcal{D}] &{=} \diag(\MC - \MC \, (\MC+\sigma^2\,\MI)^{-1}\,\MC),
\end{split}
\end{equation}
where $\MZ = (\vz_1~\vz_2~\ldots~\vz_N)^\top$ are stacked latent codes, $\MY = (\vy_1~\vy_2~\ldots~\vy_N)^\top$ are outputs from the encoder, and $\MC_{i,j} = \kappa(t_i,t_j)$ is the covariance matrix for the sequence. The posterior mean $\mathbb{E}[\vz_i \mid \mathcal{D}]$ is then passed through the decoder to get further prediction.

To obtain a covariance function that can encode the `similarity' between frames, the formulation depends on the available data: {\em (a)}~if the full pose information is available, $\mathcal{D} = \{t_i,P_i,\vy_i\}_{i=1}^N$; {\em (b)}~if only gyroscope angular rates are available, $\mathcal{D} = \{t_i,\vomega_i,\vy_i\}_{i=1}^N$; {\em (c)}~if only frame timestamps are available, $\mathcal{D} = \{t_i,\vy_i\}_{i=1}^N$. The options {\em (a)} and {\em (b)} will be covered in the next section.

For option {\em (c)}, we use a stationary Mat\'ern \cite{Rasmussen+Williams:2006} covariance function for encoding continuity and differentiability (smoothness) to the latent codes:
\begin{equation}\label{eq:timekernel}
  \kappa_\mathrm{t}(t,t') = \gamma^2\,\bigg(1 {+} \frac{\sqrt{3}\, d(t,t')}{\ell} \bigg) \\ \exp\!\bigg({-}\frac{\sqrt{3}\, d(t,t')}{\ell}\bigg),
\end{equation}
where $\gamma^2$ is a magnitude and $\ell$ a length-scale hyperparameter, and the distance metric is the time difference $d(t,t') = |t-t'|$.

In \cref{eq:gp-regression} all frames in the sequence are included at once, so the size of the covariance matrix $\MC$ grows with the number of input frames, $N$. However, for Markovian covariance functions the inference can be performed in linear time and memory complexity by converting the GP regression problem into state-space form \cite{sarkka2019applied}. This makes the information fusion approach we propose applicable for real-time and embedded systems.

\subsection{Movement-induced priors}
\label{sec:movement-induced}
Our aim is to leverage knowledge of relative movement in the disparity estimation by encoding the blunt assumption that consecutive poses should see similar scenes---the less the camera has moved and time passed, the more correlated they should be. We consider the pose prior used by Hou \etal~\cite{hou2019multi}) in monocular (multi-view) depth estimation, where the camera poses were assumed to be known. The covariance function was of form \cref{eq:timekernel} but driven by the following pose distance metric:
\begin{equation}\label{eq:pose-distance}
  d_\mathrm{pose}[P(t_i),P(t_j)] = \sqrt{\|\vp_i-\vp_j\|^2 + \frac{2}{3}\,\mathrm{tr}(\MI_3 - \MR_i^\T \MR_j)},
\end{equation}
where $\MI_3$ is a $3{\times}3$ identity matrix and `$\mathrm{tr}$' denotes the matrix trace operator. $P(t) = (\vp(t), \MR(t))$ in $\mathbb{R}^3 \times \mathrm{SO}(3)$ denotes the camera pose (rotation $\MR$ and translation $\vp$) at time instance $t$. However, this requires the pose estimation to be solved on the fly, which can be both computationally demanding, unnecessary for disparity estimation, and requiring better sensors than are available in low-cost hardware.

We now aim to relax the requirement for known poses. Consider the rotational part of \cref{eq:pose-distance} , which gives us a rotational distance metric of the form
\begin{align}\label{eq:rot-dist}
  d_\mathrm{rot}(t,t') &= \sqrt{\mathrm{tr}(\MI_3 - \MR(t)^\T\MR(t'))}.
\end{align}
Here we do not have access to $\MR$ directly, but instead we have observations of three-axis rotational rate or angular velocity of the rigid motion of the camera (\eg, observed by a gyroscope). We denote the angular velocity vector by $\vomega = ( \omega_\mathrm{x}, \omega_\mathrm{y}, \omega_\mathrm{z} )$. 

A vector $\vr$ undergoing uniform circular motion around an axis satisfies
  $\frac{\mathrm{d}\vr}{\mathrm{d}t} = \vomega \times \vr$,
where the cross-product $\vomega \times \vr$ can equivalently be expressed as an angular velocity tensor, as the cross-product matrix defined by

\begin{equation}
  [\vomega]_\times \stackrel {\rm {def}}{=}
  \begin{pmatrix}
    0  & -\omega_\mathrm{z} & \phantom{-}\omega_\mathrm{y} \\
    \phantom{-}\omega_\mathrm{z} & 0 & -\omega_\mathrm{x} \\
    -\omega_\mathrm{y} & \phantom{-}\omega_\mathrm{x} & 0
  \end{pmatrix}.
\end{equation}
By now choosing three orthonormal coordinate vectors, stacking them into $\MR$, and letting the angular velocity drive their orientation, we get the following evolution equation for the three-dimensional rotations: $\frac{\mathrm{d}\MR}{\mathrm{d}t} = [\vomega]_\times \MR$. The solution to the ODE is given by $\MR(t) = \exp(-[\vomega]_\times\,t)\,\MR(0)$, where `$\exp$' denotes the matrix exponential function.

Now consider the distance metric in \cref{eq:rot-dist} where we replace the known rotations with the corresponding rotational rate tracks, given some arbitrary initial rotation $\MR(0)$:
\begin{align}
  &\MR(t)^\T\MR(t') \nonumber \\
  &= \bigg[\int_0^t [\vomega(\tau)]_\times \MR(0) \, \mathrm{d}\tau \bigg]^\T \bigg[\int_0^{t'} [\vomega(\tau)]_\times \MR(0) \, \mathrm{d}\tau \bigg] \nonumber \\
  &= \int_t^{t'} [\vomega(\tau)]_\times \MI \, \mathrm{d}\tau.
\end{align}
Considering a piece-wise constant rotational rate (infinitesimal rotations), we can leverage the previous time-invariant solution to the ODE. For a collection of time-stamped angular velocity (gyroscope) observations $\{(t_k,\vomega_k)\}_{k=i}^j$ between time instances $t_i$ and $t_j$, we can thus write down the distance function
\begin{equation}\label{eq:fullgyrodist}\textstyle
 d_\mathrm{gyro}(t_i, t_j) = \sqrt{\mathrm{tr}(\MI_3 - \prod_{k=i+1}^{j} \exp(-[\vomega_k]_\times\,\Delta t_k))},
\end{equation}
where $\Delta t_k = t_{k}-t_{k-1}$. We disregard any possible additive or multiplicative biases in this distance metric, and simply assume the gyroscope to be suitably calibrated.

\begin{figure}[t]
  \begin{subfigure}{.3\columnwidth}
    \tikz\node[minimum width=\textwidth, minimum height=\textwidth] {\includegraphics[width=\columnwidth]{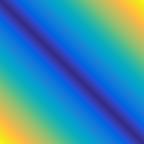}};
    \caption{$\MD_\mathrm{t}$}
    \label{fig:Dt}  
  \end{subfigure}
  \hfill
  \begin{subfigure}{.3\columnwidth}
    \tikz\node[minimum width=\textwidth, minimum height=\textwidth] {\includegraphics[width=\columnwidth]{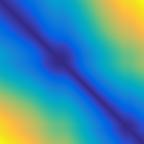}};
    \caption{$\MD_\mathrm{gyro}$}
    \label{fig:Dgyro}      
  \end{subfigure}
\hfill
  \begin{subfigure}{.3\columnwidth}
    \tikz\node[minimum width=\textwidth, minimum height=\textwidth] {\includegraphics[width=\columnwidth]{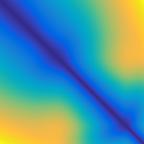}};
    \caption{$\MD_\mathrm{pose}$}
    \label{fig:Dpose}  
  \end{subfigure}
  \caption{Examples of the different distance matrix. Our Markovian gyroscope distance captures much of the same as the full pose distance, but without access to the pose information.}
  \label{fig:Dablation}

\end{figure}

\begin{table}[t]
\centering\footnotesize
\caption{Ablation experiment results on two data sets. $\kappa_\mathrm{t}$ denotes the time-decay kernel, $\kappa_\mathrm{gyro}$ the gyroscope kernel, and $\kappa_\mathrm{pose}$ the pose kernel. Results shows that the prior is always helpful, and that the gyro kernel can perform close to the full pose kernel.}
\label{tab:ablation}
\setlength{\tabcolsep}{11pt}
\begin{subfigure}{\columnwidth}
\caption{Ablation study for DispNetC-ft-seq}
\begin{tabularx}{\columnwidth}{ l | c c | c c }
\toprule
\multirow{2}{*}{Models} & \multicolumn{2}{c}{KITTI} & \multicolumn{2}{c}{ZED} \\
                        & D1-all           & Avg-all       & SSIM        & PSNR       \\ \hline
No prior             & 1.4134     & 0.6780     &      0.8770       &      32.7741    \\ 
$\kappa_\mathrm{t}$            &  1.1553&   0.6214    &       0.8755     &        32.7658    \\ 
$\kappa_\mathrm{gyro}$         &1.0846    & 0.6163     &     0.8797       &  32.8471 \\ 
$\kappa_\mathrm{t}\cdot\kappa_\mathrm{gyro}$            &  1.1339 &   0.6189    &       0.8793     &        32.8246    \\ 
$\kappa_\mathrm{pose}$        & \bf{1.0749}  &  \bf{0.6108 }   &     \bf{0.8800}       &  \bf{32.8512} \\ 
\bottomrule
\end{tabularx}
\end{subfigure}\\[1em]
\begin{subfigure}{\columnwidth}
\caption{Ablation study for PSMNet-ft}
\begin{tabularx}{\columnwidth}{ l | c c | c c }
\toprule
\multirow{2}{*}{Models} & \multicolumn{2}{c}{KITTI} & \multicolumn{2}{c}{ZED} \\
                        & D1-all           & Avg-all        & SSIM        & PSNR       \\ \hline
No prior            & 1.3878    & 1.1722     &      0.8536       &      32.2463    \\ 
$\kappa_\mathrm{t}$            &  1.3607&   1.1694   &   0.8536      &  32.2451          \\ 
$\kappa_\mathrm{gyro}$         &1.3665   & 1.1698     &   0.8537         &  32.2468 \\ 
$\kappa_\mathrm{t}\cdot\kappa_\mathrm{gyro}$ &  1.3814    & 1.1715     &   \bf{ 0.8538}      &   \bf{ 32.2487}       \\ 
$\kappa_\mathrm{pose}$ &   \bf{1.2825} &  \bf{1.1542}    &      0.8527     & 32.2352 \\ 
\bottomrule
\end{tabularx}
\end{subfigure}
  \vspace*{-1em}
\end{table}

\begin{table*}[t]
\centering\footnotesize
\caption{Experimental results on KITTI data sets benchmarking the gyroscope kernel in a GP. All results are tested with KITTI Depth val sets. Tag `-gp' means using the gyroscope-driven GP at test time, `-ft' means models are fine tuned on KITTI-2015, `-seq' means using the KITTI depth training sets as training sets. For Error rate and EPE (D1-all and Avg-all), lower number is better; for SSIM and PSNR, larger number is better.}
\label{tab:kitti}
\setlength{\tabcolsep}{9pt}
\begin{tabularx}{\textwidth}{ l | c c c | c c | c c | c c }
\toprule
\multirow{2}{*}{Model} & \multicolumn{3}{c}{Training set}            & \multicolumn{2}{c}{GP used during}     & \multicolumn{2}{c}{KITTI}  & \multicolumn{2}{c}{ZED}  \\
                       & SceneFlow & KITTI-2015      & KITTI Depth  & Training        & Testing         & D1-all      & Avg-all   & SSIM & PSNR      \\ \hline
DispNetC               & $\checkmark$   &              &              &              &              & 15.8510       & 2.3586 & 0.8446 & 32.2300    \\ 
DispNetC-gp            & $\checkmark$   &              &              &              & $\checkmark$ & 14.5620       & 1.9730 & 0.8453& 32.2356   \\
PSMNet           & $\checkmark$   &              &              &              &              & 60.0773      & 6.7620  &  0.8002 &  31.5341 \\ 
PSMNet-gp           & $\checkmark$   &              &              &              & $\checkmark$ &   60.9546   &  6.2786 &  0.7966 &  31.4693\\ 
\hline
DispNetC-ft            & $\checkmark$   & $\checkmark$ &              &              &              & 3.8739        & 0.9600  &0.8592 & 32.3660    \\ 
DispNetC-ft-gp         & $\checkmark$   & $\checkmark$ &              &              & $\checkmark$ & 3.1305      & 0.8584  & 0.8596 & 32.3726     \\
PSMNet-ft         & $\checkmark$   & $\checkmark$ &              &              &             & 1.3878        & 1.1722  & 0.8536 & 32.2463   \\
PSMNet-ft-gp         & $\checkmark$   & $\checkmark$ &              &              &      $\checkmark$       &    1.3665   & 1.1698 &  0.8537&    32.2468 \\ \hline
DispNetC-ft-seq        & $\checkmark$   &              & $\checkmark$ &  &  &  1.4134    & 0.6949 &  0.8716&  32.6382    \\ 
DispNetC-ft-seq-gp     & $\checkmark$   &              & $\checkmark$ & $\checkmark$ & $\checkmark$ & 1.0939      & 0.6155  & 0.8797 &  32.8376    \\
PSMNet-ft-seq        & $\checkmark$   &              & $\checkmark$ &  &  &  0.5391    & 0.5890 & 0.8827 &    33.0252  \\ 
PSMNet-ft-seq-gp     & $\checkmark$   &              & $\checkmark$ & $\checkmark$ & $\checkmark$ & 0.5350       & 0.5883  &   0.8829 &  33.0280      \\ \hline
\end{tabularx}
\vspace*{-1em}
\end{table*}

\cref{fig:full-dist} shows the covariance matrix based on the full cross-distance matrix corresponding to the KITTI sequence \textit{2011\_09\_26\_drive\_0005\_sync}, which consists of 144 frames with path like \cref{fig:path}. However, instead of the full cross-distance, we want to leverage the path distance in a Markovian fashion. Thus we define the cumulative pose-to-pose distance as
  $s_i = \sum_{j=1}^i d_\mathrm{gyro}(t_{j-1}, t_j)$.
We now propose a kernel (covariance function) that leverages this gyroscope distance, \cref{fig:markovian} shows the kernel based on distance $\Delta s$:
\begin{equation}\label{eq:gyrokernel}
  \kappa_\mathrm{gyro}(t_i,t_j) = \gamma^2\,\bigg(1 + \frac{\sqrt{3}\, |s_i - s_j|}{\ell} \bigg) \\ \exp\!\bigg(-\frac{\sqrt{3}\, |s_i - s_j|}{\ell}\bigg),
\end{equation}
where $\gamma^2$ is a magnitude and $\ell$ a length-scale hyperparameter. The length-scale can control the smoothness of the regression. All values of hyperparameters will be trained jointly with the encoder--decoder architectures according to the loss function (L1 loss or smooth L1 loss). \cref{fig:without,fig:with} show the sample cross-correlation of all latent encodings $\vz$ for the sequence using DispNetC and the effect after applying a GP with our gyro-kernel. We also note that because of the Markovianity, the GP can be solved in linear-time and the GP computation times can be negligible compared to evaluating the encoder/decoder.

\section{Experiments}
\label{sec:experiments}
We present extensive evaluation of the movement-induced setup. The experiments section is split into three: We provide a comprehensive ablation study in \cref{sec:ablation} for evaluating different kernels. \cref{sec:KITTI} focuses on showing the benefits of the novel and lightweight gyroscope-kernel on KITTI data which includes ground truth.  Finally, we bring over these benefits to handheld movement and lower quality inputs from a ZED camera in \cref{sec:ZED}.

We implemented our framework with the three different kernels and two architectures in PyTorch. To combine proposed prior kernels with the two representative models, for DispNetC we re-implemented it in PyTorch, using a pre-trained model on SceneFlow by \cite{tonioni2019real} and fine tuned the model on KITTI-2015 for 150 epochs (denoted `DispNetC-ft' in the tables). For PSMNet we used the original code and pre-trained models on SceneFlow and KITTI-2015 provided by the authors. As our method needs to be applied to sequences, the common benchmarks KITTI-2012, KITTI-2015, and Middlebury for disparity estimation are not ideal. Therefore we considered the KITTI depth train/validation split, using focal lengths, camera baseline, and depth labels to generate ground truth disparity maps for training and testing. We use the training set of KITTI depth to fine tune the encoder--decoder and train hyperparameters for the GP kernels jointly (denoted `DispNetC-ft-seq-gp' in the tables). And to test whether the kernel can be applied to pre-trained models directly without joint training, we directly use the learned hyperparameters (see DispNetC-gp, DispNetC-ft-gp, PSMNet-gp, PSMNet-ft-gp in the \cref{tab:kitti}). 

There are 138 sequences in the KITTI depth training set, and to jointly train the GP hyper-parameters with the encoder--decoder models, we use mini sequences which consist of three stereo pairs during training. To enable models to learn longer length-scale within fixed-length mini sequences, we randomly decide the interval length between frames. Thus there are 42{,}671 training mini-sequences in total. We trained with all kernels for 4~epochs.

\subsection{Ablation study}
\label{sec:ablation}
We conducted ablation experiments for comparing the performance of three different prior kernels: the full pose kernel $\kappa_\mathrm{pose}$ (based on \cref{eq:pose-distance}, see \cite{hou2019multi}), our new gyroscope kernel $\kappa_\mathrm{gyro}$ in \cref{eq:gyrokernel}, and a time-decay kernel $\kappa_\mathrm{t}$ in \cref{eq:timekernel}. The time-decay model can even be seen as a baseline as it can be interpreted as a trained low-pass filter. \cref{fig:Dablation} shows the computed distance matrices for the three kernel separately, where our Markovian gyroscope distance effectively captures similar pattern as the pose distance, but without access to the pose information.

\cref{tab:ablation} presents results on both KITTI Depth and ZED evaluation sets (presented in detail in \cref{sec:KITTI,sec:ZED}, respectively). In each case, the hyperparameters of the GP kernels were jointly trained with the corresponding architecture. Almost all kernels lead to better results compared to the basic architectures on KITTI, which demonstrates that introducing temporal prior knowledge among sequences is always helpful. As pose distance encodes more geometry information, the $\kappa_\mathrm{pose}$ should be the most informative prior, so it also has the best performance on both data sets for DispNetC. In contrast, the time-decay kernel $\kappa_\mathrm{t}$ is more limited as it only consider time without any geometry context. Though our proposed $\kappa_\mathrm{gyro}$ utilizes only angular rotation rates rather than the full camera pose, it still achieves comparable results with $\kappa_\mathrm{pose}$. We also test a product kernel between the gyro and time-decay which can be helpful for scenes with dynamic content and significant translation over time.

\begin{figure*}[t!]
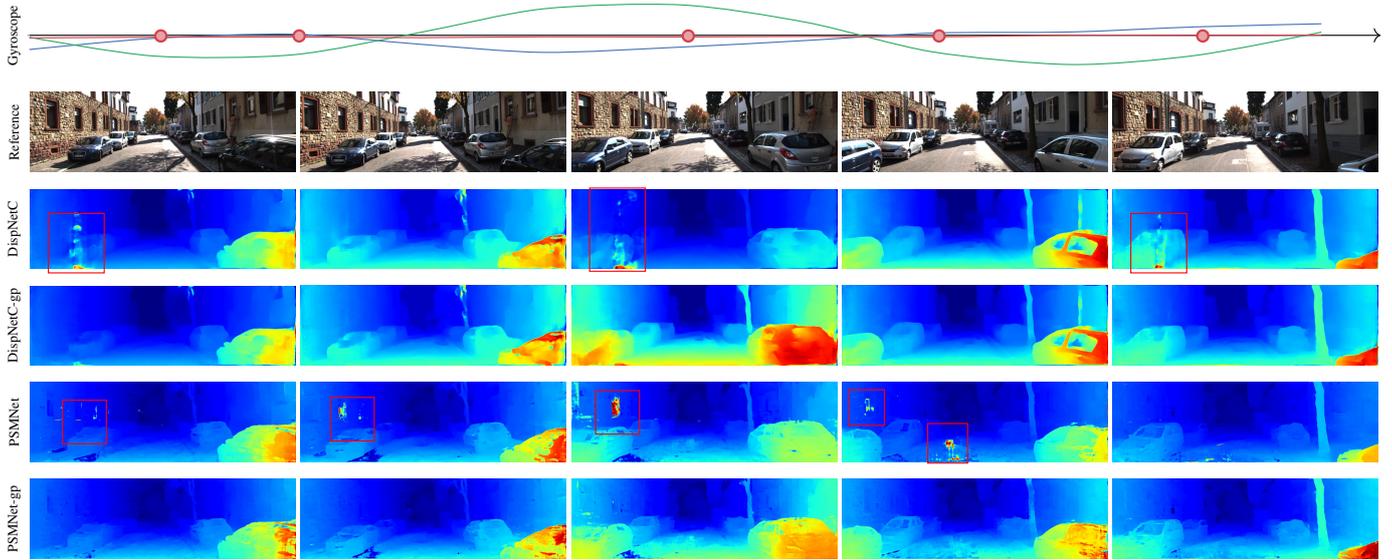

    \centering\tiny
    \setlength{\figurewidth}{.187\textwidth}
    \setlength{\figureheight}{.36\figurewidth} 
    \resizebox{\textwidth}{!}{
    \begin{tikzpicture}[inner sep=0]
      \foreach \x [count=\i] in {000, 001, 002, 003, 004}
      {

        \node[draw=white,minimum width=\figurewidth,minimum height=\figureheight,inner sep=0pt]
        (\i) at ({\figurewidth*\i},{-\figureheight*0})
        {\includegraphics[width=.98\figurewidth]{./fig/kitti/\x-ref.jpg}};

        \node[draw=white,minimum width=\figurewidth,minimum height=\figureheight,inner sep=0pt]
        (\i) at ({\figurewidth*\i},{-\figureheight*1})
{\includegraphics[width=.98\figurewidth]{./fig/kitti/\x-sf.jpg}};

        \node[draw=white,minimum width=\figurewidth,minimum height=\figureheight,inner sep=0pt]
        (\i) at ({\figurewidth*\i},{-\figureheight*2})
{\includegraphics[width=.98\figurewidth]{./fig/kitti/\x-sf-gp.jpg}};       

        \node[draw=white,minimum width=\figurewidth,minimum height=\figureheight,inner sep=0pt]
        (\i) at ({\figurewidth*\i},{-\figureheight*3})
        {\includegraphics[width=.98\figurewidth]{./fig/kitti/\x-psmnet-sf.jpg}};

        \node[draw=white,minimum width=\figurewidth,minimum height=\figureheight,inner sep=0pt]
        (\i) at ({\figurewidth*\i},{-\figureheight*4})
        {\includegraphics[width=.98\figurewidth]{./fig/kitti/\x-psmnet-sf-gp.jpg}};      
      }

      \draw[->,black] (0.5\figurewidth,\figureheight) -- (5.5\figurewidth,\figureheight);
      \node[anchor=west] at (0.5\figurewidth,\figureheight) {%
        \begin{tikzpicture}[xscale=16.2,yscale=0.4]
          
           \draw[mycolor0] plot [smooth] coordinates{ (0.000,-0.431) (0.097,-0.076) (0.204,0.055) (0.301,-0.257) (0.398,-0.519) (0.505,-0.351) (0.602,-0.142) (0.699,0.090) (0.806,0.125) (0.903,0.286) (1.000,0.383) };
           \draw[mycolor1] plot [smooth] coordinates{ (0.000,-0.039) (0.097,-0.635) (0.204,-0.597) (0.301,0.096) (0.398,0.855) (0.505,0.979) (0.602,0.425) (0.699,-0.502) (0.806,-0.903) (0.903,-0.613) (1.000,0.120) };
           \draw[mycolor2] plot [smooth] coordinates{ (0.000,-0.065) (0.097,-0.040) (0.204,-0.034) (0.301,-0.056) (0.398,-0.028) (0.505,-0.034) (0.602,-0.015) (0.699,-0.031) (0.806,0.024) (0.903,0.018) (1.000,0.030) };

           \node[shape=circle,draw=mycolor2,fill=mycolor2!50,thick,minimum width=4pt,inner sep=0] at (0.097,0.000) {};
           \node[shape=circle,draw=mycolor2,fill=mycolor2!50,thick,minimum width=4pt,inner sep=0] at (0.204,0.000) {};
           \node[shape=circle,draw=mycolor2,fill=mycolor2!50,thick,minimum width=4pt,inner sep=0] at (0.505,0.000) {};
           \node[shape=circle,draw=mycolor2,fill=mycolor2!50,thick,minimum width=4pt,inner sep=0] at (0.699,0.000) {};
           \node[shape=circle,draw=mycolor2,fill=mycolor2!50,thick,minimum width=4pt,inner sep=0] at (0.903,0.000) {};
           
           \draw[white] (0,-1) -- (0,1);
                              
        \end{tikzpicture}      
      };

      \node[rotate=90] at ({0.45*\figurewidth},{1*\figureheight}) {Gyroscope};      
      \node[rotate=90] at ({0.45*\figurewidth},{-0*\figureheight}) {Reference};
      \node[rotate=90,text width=1.5cm,align=center] at ({0.45*\figurewidth},{-1*\figureheight}) {DispNetC};
      \node[rotate=90,text width=1.5cm,align=center] at ({0.45*\figurewidth},{-2*\figureheight}) {DispNetC-gp};
      \node[rotate=90,text width=1.5cm,align=center] at ({0.45*\figurewidth},{-3*\figureheight}) {PSMNet};
      \node[rotate=90,text width=1.5cm,align=center] at ({0.45*\figurewidth},{-4*\figureheight}) {PSMNet-gp};

      \node[shape=rectangle,minimum height=7.5mm, minimum width=7mm,draw=red] at (0.68\figurewidth,-1.15\figureheight) {};
      \node[shape=rectangle,minimum height=10.5mm, minimum width=7mm,draw=red] at (2.68\figurewidth,-1.01\figureheight) {};
      \node[shape=rectangle,minimum height=7.5mm, minimum width=7mm,draw=red] at (4.68\figurewidth,-1.15\figureheight) {};

      \node[shape=rectangle,minimum height=5.5mm, minimum width=5.5mm,draw=red] at (0.71\figurewidth,-3\figureheight) {};
     \node[shape=rectangle,minimum height=5.5mm, minimum width=5.5mm,draw=red] at (1.7\figurewidth,-2.97\figureheight) {};
      \node[shape=rectangle,minimum height=5.5mm, minimum width=5.5mm,draw=red] at (2.68\figurewidth,-2.9\figureheight) {};
      \node[shape=rectangle,minimum height=4.5mm, minimum width=4.5mm,draw=red] at (3.6\figurewidth,-2.85\figureheight) {};
     \node[shape=rectangle,minimum height=5mm, minimum width=5mm,draw=red] at (3.9\figurewidth,-3.22\figureheight) {};
      
    \end{tikzpicture}}
  \caption{Example sequence from KITTI {\it 2011\_09\_26\_drive\_0095\_sync}. Both the DispNetC and PSMNet pre-trained from synthetic data show artifacts, and our prior helps to alleviate them. The accompanying faster-sampled gyroscope data from the KITTI data set is visualized on the top with the circles marking the frame timings of the these five frames. For comprehensive quantitative results, see \cref{tab:kitti,tab:ablation}.}
  \label{fig:kitti}
  \vspace*{-1em}
\end{figure*}

\begin{figure}[t!]
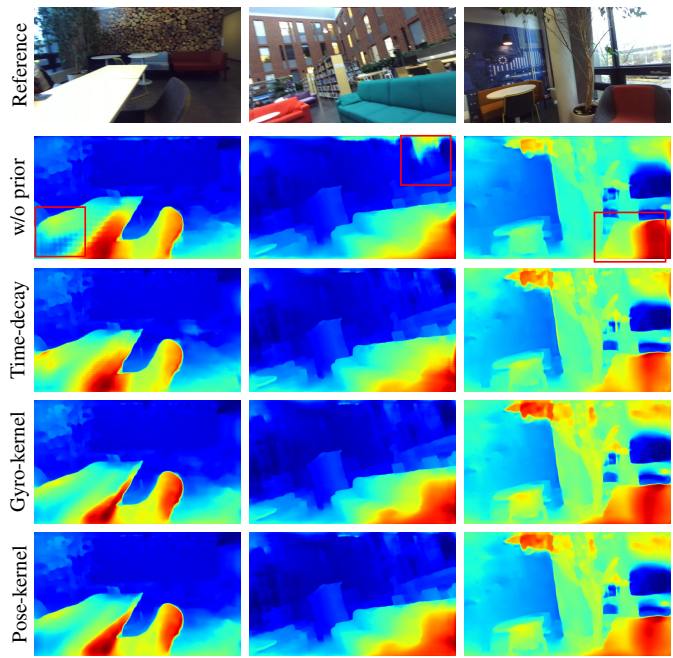

   \centering\tiny
   \setlength{\figurewidth}{0.1\textwidth}
   \setlength{\figureheight}{0.62\figurewidth} 
    \resizebox{\columnwidth}{!}{
    \begin{tikzpicture}[inner sep=0]
      \foreach \x [count=\i] in {000, 001, 002}
      {

        \node[draw=white,minimum width=\figurewidth,minimum height=\figureheight,inner sep=0pt]
        (\i) at ({\figurewidth*\i},{-\figureheight*0})
        {\includegraphics[width=.96\figurewidth]{./fig/ZED/\x-ref.jpg}};
       
        \node[draw=white,minimum width=\figurewidth,minimum height=\figureheight,inner sep=0pt]
        (\i) at ({\figurewidth*\i},{-\figureheight*1})
{\includegraphics[width=.96\figurewidth]{./fig/ZED/\x-wo.jpg}};
        
        \node[draw=white,minimum width=\figurewidth,minimum height=\figureheight,inner sep=0pt]
        (\i) at ({\figurewidth*\i},{-\figureheight*2})
{\includegraphics[width=.96\figurewidth]{./fig/ZED/\x-Kt.jpg}};       
        
        \node[draw=white,minimum width=\figurewidth,minimum height=\figureheight,inner sep=0pt]
        (\i) at ({\figurewidth*\i},{-\figureheight*3})
        {\includegraphics[width=.96\figurewidth]{./fig/ZED/\x-Kgyro.jpg}};
        
        \node[draw=white,minimum width=\figurewidth,minimum height=\figureheight,inner sep=0pt]
        (\i) at ({\figurewidth*\i},{-\figureheight*4})
        {\includegraphics[width=.96\figurewidth]{./fig/ZED/\x-Kpose.jpg}};      
      }

      \node[rotate=90] at ({0.45*\figurewidth},{-0*\figureheight}) {Reference};
      \node[rotate=90,text width=1.5cm,align=center] at ({0.45*\figurewidth},{-1*\figureheight}) {w/o prior};
      \node[rotate=90,text width=1.5cm,align=center] at ({0.45*\figurewidth},{-2*\figureheight}) {Time-decay};
      \node[rotate=90,text width=1.5cm,align=center] at ({0.45*\figurewidth},{-3*\figureheight}) {Gyro-kernel};
      \node[rotate=90,text width=1.5cm,align=center] at ({0.45*\figurewidth},{-4*\figureheight}) {Pose-kernel};

      \node[shape=rectangle,minimum height=4.2mm, minimum width=4.2mm,draw=red] at (0.64\figurewidth,-1.26\figureheight) {};

      \node[shape=rectangle,minimum height=4.2mm, minimum width=4.2mm,draw=red] at (2.34\figurewidth,-.72\figureheight) {};

      \node[shape=rectangle,minimum height=4.2mm, minimum width=6mm,draw=red] at (3.29\figurewidth,-1.3\figureheight) {};

    \end{tikzpicture}}
  \caption{Example frames from sequences collected with a ZED camera and results with DispNetC. Using priors helps in texture-less regions like the table and the wall (see red boxes). Here the full pose-kernel can be seen as close to ground truth, and the gyro-kernel performance matches it well, even if the requirements for it are more lightweight.}
  \label{fig:zed}
  \vspace*{-1em}
\end{figure}

\subsection{Evaluation on KITTI}
\label{sec:KITTI}
For evaluation on KITTI, there are 13 sequences which include 3426 left--right pairs. We use both Average end-point-error in total (Avg-all) and percentage of outliers (estimation error is larger than 3~px and larger than 5\% of the ground truth disparity at this pixel) averaged over all ground truth pixels (D1-all), to measure the performance. 

\cref{tab:kitti} shows the performance on the KITTI depth validation data set. All `-gp' models in the table correspond to the proposed novel gyroscope kernel \cref{eq:gyrokernel}. All baseline models that were only pre-trained with synthetic data suffer from domain shift, and the artifacts lead to a high error rate. According to the EPE metric, introducing our prior information makes improvement for both pre-trained DispNetC and pre-trained PSMNet. For PSMNet, as the baseline model have particular high error rate, there are several sequences have similar artifacts every frame, so our fusion scheme will lead to higher error rate in this case. But if the artifacts only appear in several frames in the whole sequences, our method can fix artifacts automatically. \cref{fig:kitti} shows how our method alleviates the artifacts. Models with tag `-ft' are fine tuned on KITTI-2015; the results show that after fine tuning all models adapt to the new domain, while our method can still improve the performance further based on that. The tag `-ft-seq' corresponds to models with training from pre-trained model with KITTI depth sequences. Because of the larger size of training samples, the resulting baseline models get the best performance, and even the room for improvement becomes limited, the prior can still refine the results. 

During inference, the frame rate for the PyTorch-implemented DispNetC is 154.3~fps, and after adding the GP prior the frame rate is 50.6~fps. The frame rate for PSMNet is 3~fps, and after using the GP prior the frame rate is 3.1~fps. The additional runtime introduced by the GP regression per frame is  $\sim$0.01~s, which shows that the method is time-efficient.

\subsection{Evaluation on ZED}
\label{sec:ZED}
As most sequences in KITTI are moving straight at near constant speed and scenes share similar structure, we also sought more varied motion and environments for testing. Due to lack of suitable public data sets, we collected five indoor sequences by using a hand-held ZED stereo camera (\url{https://www.stereolabs.com}), which provides both left/right image pairs and tracking information. The resolution of collected pairs is $1280{\times}720$ and the capture rate is 10~fps. There are 2194 test pairs in total. The full-pose information is also provided by the sensor,
and to evaluate our proposed gyroscope-kernel, we derivate the angular velocities by converting the quaternion sequences.
This data illustrates the kind of challenging motion and varied use cases that, \eg, smartphone data would feature.

However, since there is no depth sensor on the ZED camera, there is no ground truth depth/disparity map provided. In that case, to evaluate results, we use predicted disparity maps and the right images to synthesize left images,  and then use structural similarity (SSIM) and peak signal-to-noise ratio (PSNR) to measure the similarity between warped left images and original left images. For these metrics, larger numbers mean higher similarity, which indicates higher quality of predictions. Specifically, the SSIM is given by
\begin{equation}
  \mathrm{SSIM}(x,y) =\frac{(2\,\mu_\mathrm{x}\mu_\mathrm{y}+c_1)(2\,\sigma_\mathrm{xy}+c_2)}{(\mu_\mathrm{x}^2+\mu_\mathrm{y}^2+c_1)(\sigma_\mathrm{x}^2+\sigma_\mathrm{y}^2+c_2)},
\end{equation}
where for each input windows $x$ and $y$, the $\mu_\mathrm{x}$ and $\mu_\mathrm{y}$ are the average and $\sigma_\mathrm{x}^2$ and $\sigma_\mathrm{y}^2$ are variances over windows $x$ and $y$ correspondingly, $\sigma_\mathrm{xy}$ denotes the cross-covariance. We use constants $c_1=0.01^2$ and $c_2=0.03^2$. The other metric is
  $\mathrm{PSNR}(x,y) = 20\log_{10}\left(\frac{\mathrm{MAXI}}{\sqrt{\mathrm{MSE}}}\right)$,
where $\mathrm{MAXI}$ is the maximum possible pixel value of the image (here is 255) and $\mathrm{MSE}$ is the mean squared error over inputs $x$ and $y$. 

In \cref{tab:kitti}, almost all models present similar trends on both KITTI and ZED data sets. For DispNet the fusion always helps. For baseline models that suffer from domain shifting, it is difficult to boost the performance by only fusing the latent code.  After jointly training the GP kernel and the encoder--decoder, though the training set is still different from evaluation domain, the results get better.  And we have more results in \cref{sec:ablation} to show that introducing prior can be beneficial for PSMNet without re-training.

\section{Discussion and conclusion}
\label{sec:discussion}
In the paper, we showed that introducing temporal priors for state-of-the-art deep stereo methods can improve performance, and proposed a novel movement-induced kernel that only needs angular rates rather than full camera pose. Our latent-space fusion strategy can be applied to almost all existing learning-based stereo matching methods to regularize the cost volume. 
Thus this work provides a complementary track for improving deep stereo estimation. In the experiments, we demonstrated that the technique can both be incorporated with pre-trained model directly or can be used in fine-tuning pre-trained models and GP hyperparameters jointly. Our qualitative results show that leveraging movement-induced priors can help alleviate artifacts caused by domain shift via enforcing temporal consistency.

As the method mainly aims to solve the inconsistency among frames in sequences, one limitation is that if the baseline predictions on all frames are wrong, the prior cannot boost performance as they are consistently bad already. Another limitation is that we only consider the movement of cameras, and disregards dynamic objects. Dealing with those dynamic objects can be a direction for future work.

The codes will be made available at \url{https://github.com/AaltoVision/movement-induced-prior}.

\section*{Acknowledgments}
The authors acknowledge the computational resources provided by the Aalto Science-IT project. This research was supported by the Academy of Finland grants 308640, 324345 and 309902.

{\small
\bibliographystyle{IEEEtran}
\bibliography{bibliography}
}

\clearpage
\twocolumn[\vspace*{2em}\centering\Large\bf Supplementary Material for\\ Movement-induced  Priors for Deep Stereo\vspace*{2em}]

\appendix

This is the supplementary material for `Movement-induced Priors for Deep Stereo'. The supplementary material comprises this appendix and an associated video supplement.

\section{Architecture details }
\label{app:architecture}
We present the relevant parts of the architecture of the models considered as encoder--decoders in the main paper: DispNetC \cite{mayer2016large} and PSMNet\cite{chang2018pyramid}.

\subsection{Architecture details of DispNetC}
Following \cite{mayer2016large} and \cite{dosovitskiy2015flownet}, the architecture details of the DispNetC model used in the first encoder--decoder setup in the main paper are described in \cref{tb:arch-dispnetc}. We only summarize the model in form of a table as it directly conforms the setup in \cref{fig:architecture}.

For the 1D correlation layer, the neighborhood size is $D=2d+1$, while the maximum displacement is $d=40$. The correlation of two pixel at $\vx_1$ in the left map and $\vx_2$ in the right map is computed by the scalar product of two feature vectors $c(\mbf{x}_1, \mbf{x}_2) = \left \langle \mbf{f}_1(\mbf{x}_1), \mbf{f}_2(\mbf{x}_2)  \right \rangle$. 

We split the original DispNetC into an encoder and decoder. The encoder
part consists of convolutions \texttt{conv1} to \texttt{conv6b}, and in the decoder part upconvolutions (\texttt{upconvN}), convolutions (\texttt{iconvN}, \texttt{prN}) are alternating. Our GP kernel will only be applied to the output of the encder \texttt{conv6b}, while other parts remain unchanged.

\begin{table}[!t]
\small
\caption{Architecture details of DispNetC. \texttt{prN} means prediction layers that generate prediction at different scale. \texttt{+} means concatenation.The input left and right images are processed separately up to \texttt{conv2} and then the 1D correlation layer compute the correlation score via scalar product. We split the original rchitecture into the encoder and the decoder and our GP kernel will be applied to the output of the encoder \texttt{conv6b}.}
\label{tb:arch-dispnetc}

\begin{tabular}{|c|c|c|c|c|}
\hline
Name      & Kernel       & Str. & Ch I/O        & Input              \\ \hline
conv1L    & 7$\times$7 & 2      & 3/64          & left image         \\ 
conv1R    & 7$\times$7 & 2      & 3/64          & right image        \\ 
conv2L    & 5$\times$5 & 2      & 64/128        & conv1L             \\ 
conv2R    & 5$\times$5 & 2      & 64/128        & conv1R             \\ \hline
corr      & ---            & ---      & ---             & conv2L, conv2R     \\ 
redir     & 1$\times$1 & 1      & 128/64        & conv2L             \\ \hline
conv3a    & 5$\times$5 & 2      & 64+D/256 & redir +corr        \\ 
conv3b    & 3$\times$3 & 1      & 256/256       & conv3a             \\ 
conv4a    & 3$\times$3 & 2      & 256/512       & conv3b             \\
conv4b    & 3$\times$3 & 1      & 512/512       & conv4a             \\ 
conv5a    & 3$\times$3 & 2      & 512/512       & conv4b             \\ 
conv5b    & 3$\times$3 & 1      & 512/512       & conv5a             \\ 
conv6a    & 3$\times$3 & 2      & 512/1024      & conv5b             \\ 
conv6b    & 3$\times$3 & 1      & 512/1024      & conv6a             \\ \hline  \hline
pr6 & 3$\times$3 & 1      & 1024/1        & conv6b             \\ \hline
upconv5   & 4$\times$4 & 2      & 1024/512      & conv6b             \\
iconv5    & 3$\times$3 & 1      & 1024/512      & upconv5+pr6+conv5b \\ 
pr5 & 3$\times$3 & 1      & 512/1         & iconv5             \\ \hline
upconv4   & 4$\times$4 & 2      & 512/256       & iconv5             \\ 
iconv4    & 3$\times$3 & 1      & 769/256       & upconv4+pr5+conv4b \\ 
pr4 & 3$\times$3 & 1      & 256/1         & iconv4             \\  \hline
upconv3   & 4$\times$4 & 2      & 256/128       & iconv4             \\ 
iconv3    & 3$\times$3 & 1      & 385/128       & upconv3+pr4+conv3b \\ 
pr3 & 3$\times$3 & 1      & 128/1         & iconv3             \\ \hline
upconv2   & 4$\times$4 & 2      & 128/64        & iconv3             \\
iconv2    & 3$\times$3 & 1      & 193/64        & upconv2+pr3+conv2L  \\ 
pr2 & 3$\times$3 & 1      & 64/1          & iconv2             \\ \hline
upconv1   & 4$\times$4 & 2      & 64/32         & iconv2             \\ 
iconv1    & 3$\times$3 & 1      & 97/32         & upconv1+pr2+conv1L  \\ 
pr1 & 3$\times$3 & 1      & 32/1          & iconv1             \\ \hline
\end{tabular}
\end{table}

\subsection{Modified architecture of PSMNet}
The second model architecture considered in the main paper was PSMNet~\cite{chang2018pyramid}. The setup for PSMNet is more involved as the re-casting of the model into an encoder--decoder formulation (refer to \cref{fig:architecture} for the basic setup of our model) leaves more freedom.

\begin{figure*}
 \resizebox{\textwidth}{!}{%
 \begin{tikzpicture}[xscale=2.75]
 
 \newcommand{\costvol}{\tikz\draw (0,0) -- (1,0) -- (1,1) -- (0,1) -- (0,0) (1,0) -- (1.25,.5) -- (1.25,1.25) -- (0.25,1.25) -- (0,1) (1,1) -- (1.25,1.25);}
 
 \tikzstyle{block} = [rounded corners=2pt,minimum width=17mm,minimum height=7mm,inner sep=3pt,draw=black,fill=white]
 
 \node[] (I1) at (0,1) {\includegraphics[width=2.5cm]{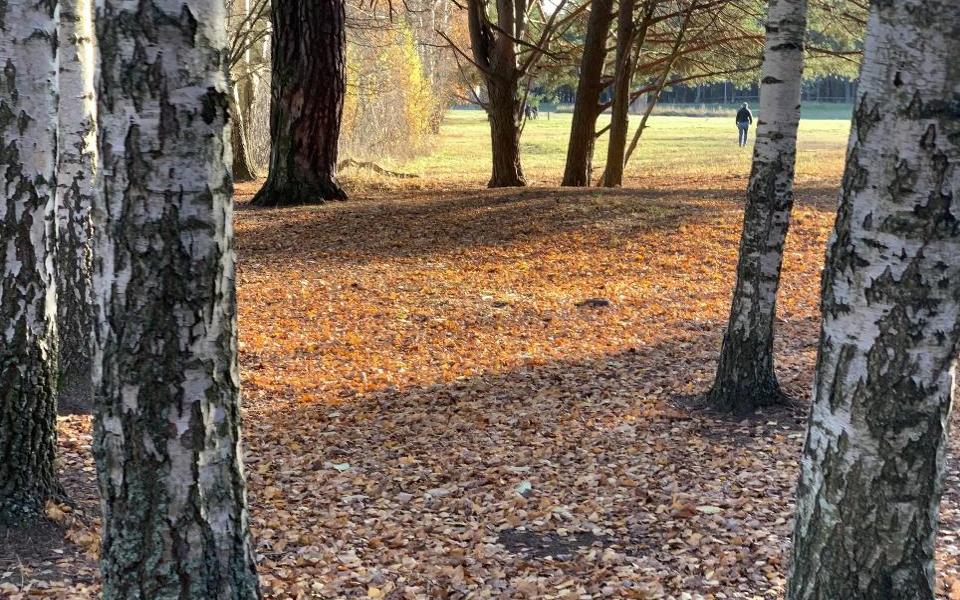}};
 \node[] (I2) at (0,-1) {\includegraphics[width=2.5cm]{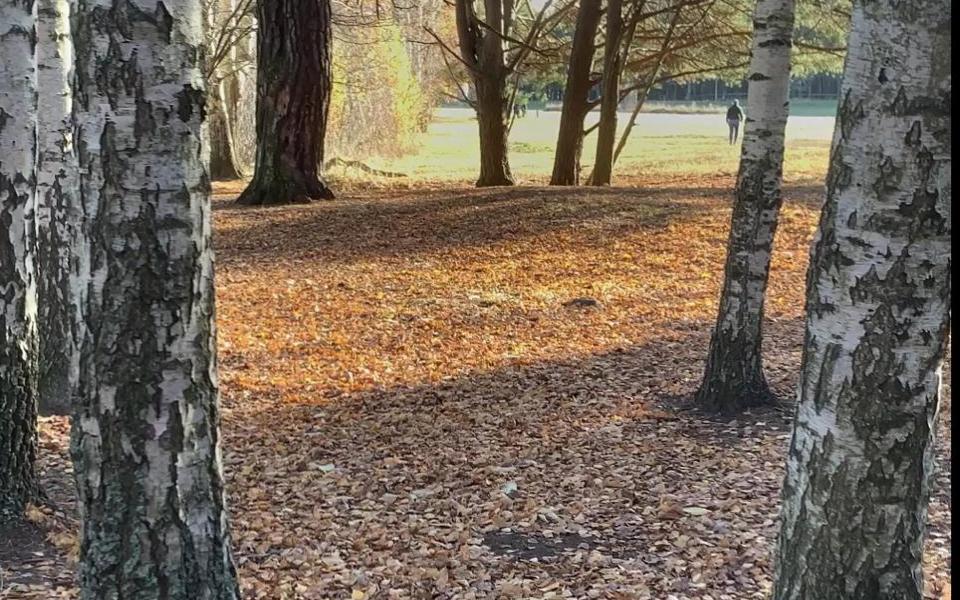}}; 
 \node[above of=I1] {Left image};
 \node[below of=I2] {Right image}; 
 
 \node[block] (CNN1) at (1,1) {CNN};
 \node[block] (CNN2) at (1,-1) {CNN}; 
 \node (W1) at (1,0) {weight sharing};
 \draw[-latex] (W1) -> (CNN1);
 \draw[-latex'] (W1) -> (CNN2);
 
 \node[block] (SPP1) at (2,1) {SPP module};
 \node[block] (SPP2) at (2,-1) {SPP module}; 
 \node (W2) at (2,0) {weight sharing};
 \draw[-latex] (W2) -> (SPP1);
 \draw[-latex'] (W2) -> (SPP2);
 
 \node[block] (CON1) at (3,1) {CONV};
 \node[block] (CON2) at (3,-1) {CONV};
 \node (W3) at (3,0) {weight sharing};
 \draw[-latex] (W3) -> (CON1);
 \draw[-latex'] (W3) -> (CON2); 
 
 \node[] (COST) at (4,0) {\costvol};
 \node[below of=COST] {Cost volume};

 \node[block] (CNN3D) at (5,0) {3D CNN};
 
 \node[rotate=90,block] (UP) at (5.75,0) {upsampling};
 \node[rotate=90,block] (REG) at (6.25,0) {regression};
 
 \node[] (FIN) at (7.25,0) {\includegraphics[width=2.5cm]{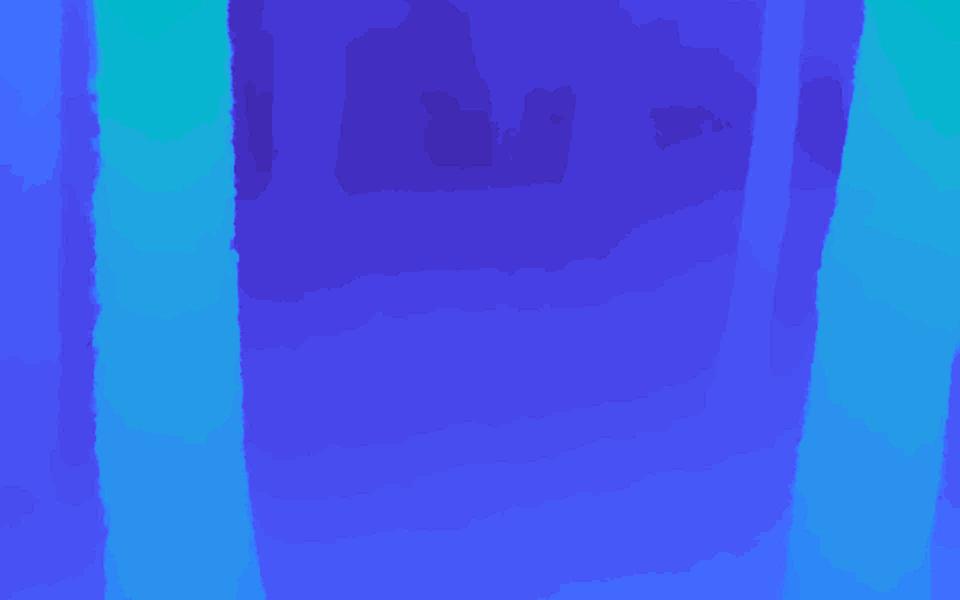}};
 \node[below of=FIN] {Final prediction};

 \draw[-latex] (I1) -> (CNN1);
 \draw[-latex] (I2) -> (CNN2);
 \draw[-latex] (CNN1) -> (SPP1);
 \draw[-latex] (CNN2) -> (SPP2);
 \draw[-latex] (SPP1) -> (CON1);
 \draw[-latex] (SPP2) -> (CON2);
 \draw[-latex] (CON1) -> (COST);
 \draw[-latex] (CON2) -> (COST);
 \draw[-latex] (COST) -> (CNN3D);
 \draw[-latex] (CNN3D) -> (UP);
 \draw[-latex] (UP) -> (REG);
 \draw[-latex] (REG) -> (FIN);

 \draw[mycolor0,dashed,rounded corners=2pt,thick] (1.5,-1.75) rectangle ++(1,3.5);
 \draw[mycolor1,dashed,rounded corners=2pt,thick] (4.5,-1.25) rectangle ++(2,2.5);

  \node[draw=mycolor0,dashed,rounded corners=2pt,thick,text width=7cm,minimum height=6cm,line width=2pt,inner sep=1em,align=center,anchor=north] (SPPB) at (0.75,-4) {\includegraphics[width=6cm]{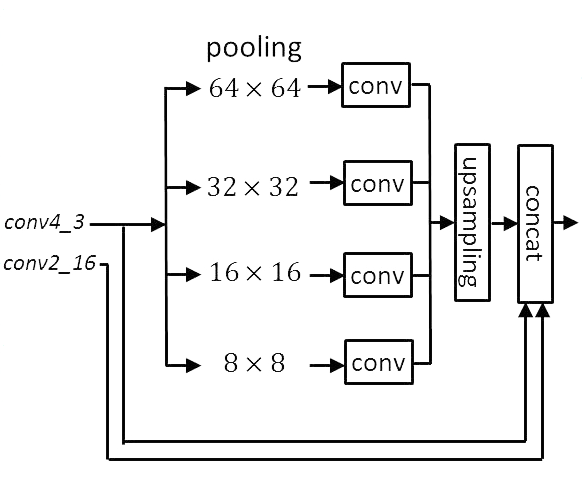}}; 

  \node[draw=mycolor1,dashed,rounded corners=2pt,thick,text width=14cm,minimum height=6cm,line width=2pt,inner sep=1em,align=center,anchor=north] (CNNB) at (5,-4) {\includegraphics[width=14cm]{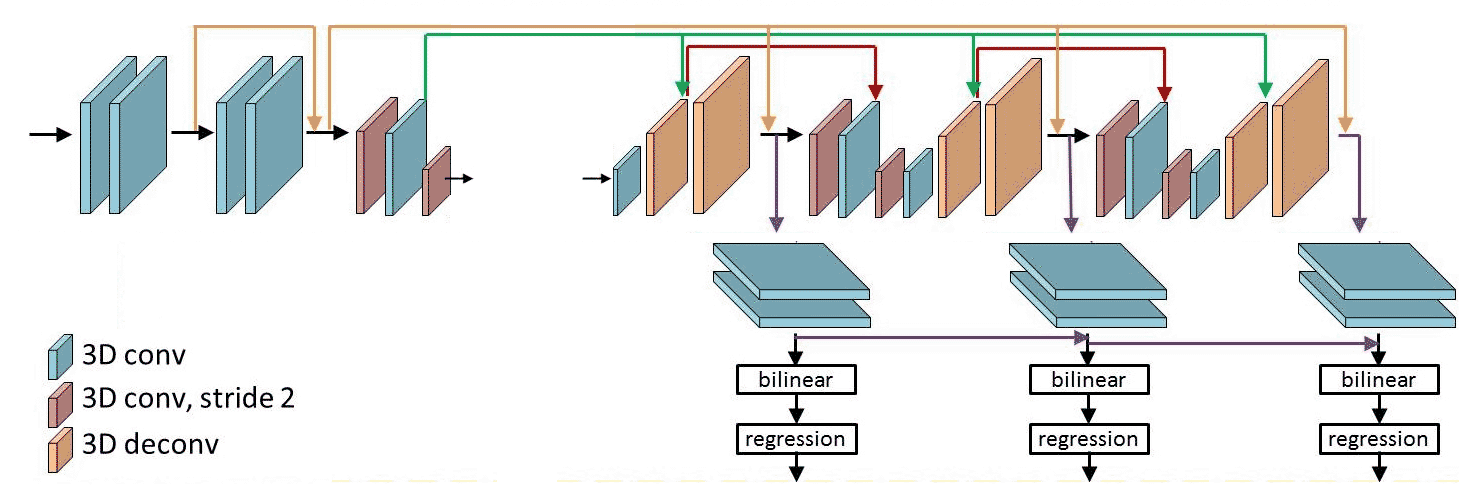}};

  \node[above of=SPPB,yshift=2.4cm] {\textbf{Spatial Pyramidal Pooling Module}};
  \node[above of=CNNB,yshift=2.4cm] {\textbf{Modified stacked hourglass}};

  \node[shape=circle,draw=mycolor0,fill=mycolor0!50,thick,minimum width=2em] at (4.3,-6.35) {$\vz_i$};
  \node[text width=2cm, align=center] at (4.3,-7.5) {Latent space encoding};

  \draw[-latex,mycolor0,thick] (2,-1.75) to[out=270, in=90] (0.75,-3.25);
  \draw[-latex,mycolor1,thick] (5.5,-1.25) to[out=270, in=90] (5,-3.25);
  
 \end{tikzpicture}}
 \vspace*{1em}
 \caption{Architecture details adopted from the original PSMNet~\cite{chang2018pyramid} (check paper for additional details) with our modfied Stacked hourglass architecture. Only the first encoder--decoder in the Stacked hourglass module is affected. All outputs of the first encoder among the sequence will be used to compute $\vz_i$ via GP regression, and the $\vz_i$ will feed the following decoder (\cf, \cref{fig:architecture}).}
\label{fig:psmnet}
\end{figure*}
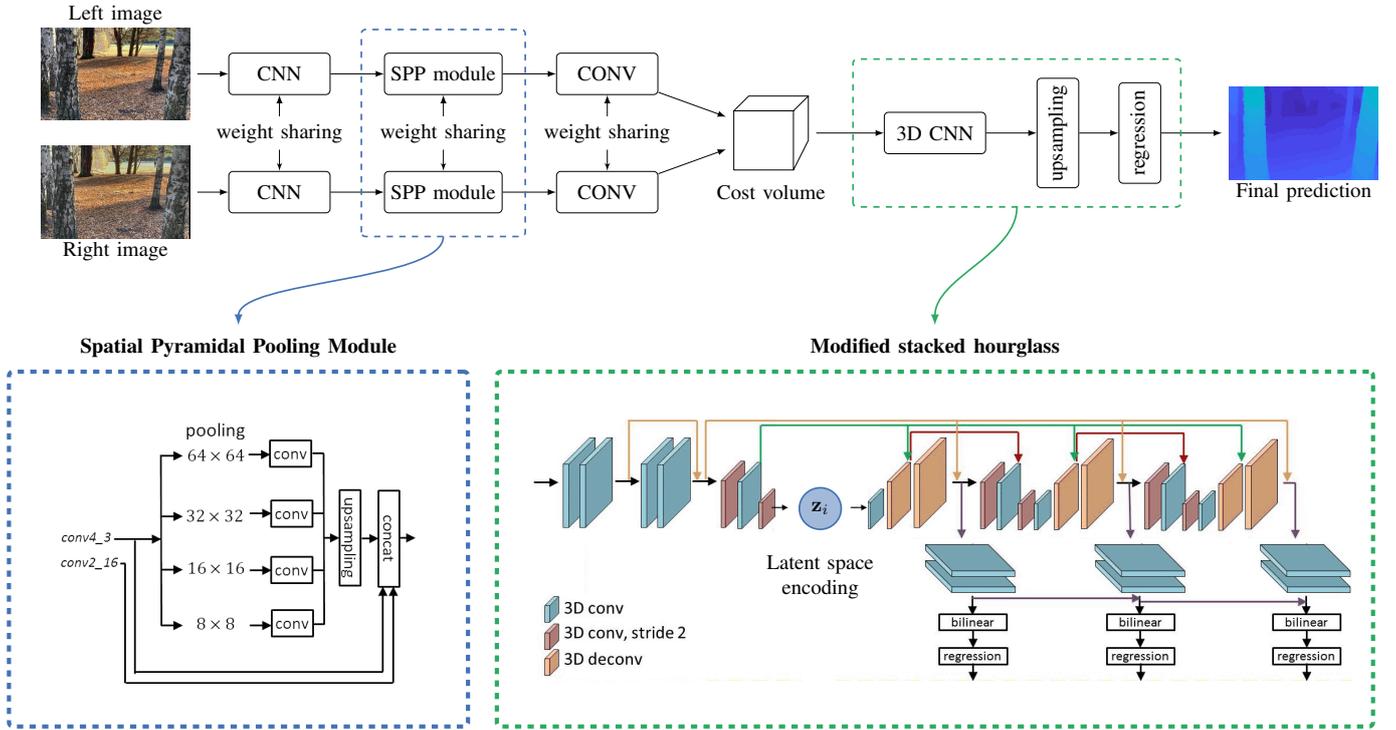

\cref{fig:psmnet} shows both the architecture of the original PSMNet with our modifications. We have only revised the stacked hourglass module, and the prior is assigned to the first enoder--decoder of the three-part hourglass chain. As shown in \cref{fig:psmnet}, all outputs of the first encoder among the sequence will be used to compute $\vz_i$ via GP regression, and the $\vz_i$ will feed the following decoder.

\subsection{Conversion of orientations to angular rates}
\label{app:quat-to-gyro}
In order to avoid gimbal-lock and to streamline handling of rigid motions, quaternions provide a commonly used representation of rotations in three-dimensional space. In the experiments in the main paper, we leveraged pose information provided by the ZED camera for evaluating the pose-kernel. However, the proposed gyroscope-kernel uses angular velocities (what a three-axis gyroscope would provide) and in order to run direct comparisons between the methods, we wanted to backtrack the genrating angular velocities from the ZED camera poses.

For reference, we provide the following short derivation of the conversion of orientation unit quaternion sequences to angular velocities.

We represent rotations as unit quaternions that can be represented as 4-dimensinal vectors $\vq(t) \in \R^4$ such that $\vq = (q_\mathrm{w}, q_\mathrm{x}, q_\mathrm{y}, q_\mathrm{z})$ and $\|\vq\|=1$. The time derivative will be 
\begin{equation}
  \dot{\vq}(t) = \frac{\mathrm{d}}{\mathrm{d}t} \vq(t) = \lim_{\delta \to 0}\frac{\vq(t+\delta) - \vq(t)}{\delta}.
\end{equation}
We now denote the angular velocity as $\vomega(t) \in \R^3$, which following the logic in also presented in the main paper, drives the differential rotation
\begin{equation}
  \dot{\vq}(t) = \frac{1}{2}\,\vq(t)\odot\begin{pmatrix} 0 \\ \vomega(t)  \end{pmatrix}.
\end{equation}
Solving for unknown angular velocity gives
\begin{equation}
  \vomega(t) = \mathrm{Im}[2\,\vq^*(t) \odot \dot{\vq}(t)],
\end{equation}
where $\vq^*$ is quaternionic conjugation of unit quaternion $\vq$, the operator $\odot$ is the quaternion prodcut, and $\mathrm{Im}$ means extracting the imaginary part of the quaternion ($(q_\mathrm{x}, q_\mathrm{y}, q_\mathrm{z}) = \mathrm{Im}[(q_\mathrm{w}, q_\mathrm{x}, q_\mathrm{y}, q_\mathrm{z})]$). We then use finite differences of observed pose rotations in the ZED data to calculae the angular velocities.

\end{document}